%% file: main.tex
\documentclass[11pt]{article}

\usepackage[final]{acl}

\usepackage{times}
\usepackage{latexsym}

\usepackage[T1]{fontenc}
\usepackage[utf8]{inputenc}

\usepackage{microtype}

\usepackage{inconsolata}

\usepackage{graphicx}
\input{notations}

\usepackage{amsmath}
\usepackage{amssymb}

\usepackage{booktabs}
\usepackage{microtype}
\usepackage{graphicx}
\usepackage{cuted} % permits fixed full-width material in a two-column page
\usepackage{afterpage}
\usepackage{subcaption}
\usepackage{booktabs} % for professional tables

\usepackage{hyperref}

\usepackage{pgfplots}
\pgfplotsset{compat=1.18}
\usepgfplotslibrary{fillbetween}

\usepackage{listings}
\usepackage[most]{tcolorbox}
\usepackage{xcolor}

\newtcolorbox{goldbox}{
  breakable,
  colback=yellow!6,
  colframe=yellow!45!black,
  boxrule=0.8pt,
  arc=2mm,
  left=1.5mm,
  right=1.5mm,
  top=1mm,
  bottom=1mm
}

\newtcolorbox{redbox}[1]{
  breakable,
  colback=red!4,
  colframe=red!65!black,
  title=#1,
  fonttitle=\bfseries,
  boxrule=0.8pt,
  arc=2mm,
  left=1.5mm,
  right=1.5mm,
  top=1mm,
  bottom=1mm
}

\newtcolorbox{greenbox}[1]{
  breakable,
  colback=green!4,
  colframe=green!50!black,
  title=#1,
  fonttitle=\bfseries,
  boxrule=0.8pt,
  arc=2mm,
  left=1.5mm,
  right=1.5mm,
  top=1mm,
  bottom=1mm
}

\usepackage[capitalize,noabbrev]{cleveref}

\newcommand{\inputs}{\mathbf{s}}
\newcommand{\data}{\mathcal{D}}

\newcommand{\corrupts}{\tilde{s}}
\newcommand{\corruptinputs}{\tilde{\mathbf{s}}}
\newcommand{\corruptD}{\mathcal{Q}}
\newcommand{\ARobj}{\mathcal{L}_{\text{NTP}}}
\newcommand{\CATobj}{\mathcal{L}_{\text{CAT}}}
\usepackage[textsize=tiny]{todonotes}

\title{\maintitle}

\author{
  \normalsize\textbf{Meghanadh Pulivarthi\footnotemark[1] \quad Kushagra Bhushan\footnotemark[1] \quad Vineet Kumar\footnotemark[2] \quad Gaurav Pandey} \\
  \normalsize\textbf{Jaydeep Sen \quad Dinesh Raghu \quad Sachindra Joshi \quad Yatin Nandwani} \\
  \normalsize IBM \\
  \normalsize\texttt{\{Meghanadh.Pulivarthi1, kushagrabhushan, Yatin.Nandwani\}@ibm.com} \\
  \normalsize\texttt{\{gpandey1, jaydesen, diraghu1, jsachind\}@in.ibm.com} \\
  \normalsize\texttt{vineet.mundhra@gmail.com}
}

\begin{document}
\maketitle

% put these right after \maketitle
\renewcommand{\thefootnote}{\fnsymbol{footnote}}
\footnotetext[1]{Equal contribution.}
\footnotetext[2]{Work done while at IBM. Currently at Amazon Books Science.}
\renewcommand{\thefootnote}{\arabic{footnote}}  % restore for body footnotes

\input{sections/abstract}
\input{sections/introduction}
\input{sections/related_works_main}
\input{sections/method}
\input{sections/experiments}
\input{sections/experimental_results}
\input{sections/conclusion}

\section*{Limitations}
A limitation of \ourmethodshort\ is that, while corruption-based augmentation reduces the need for expensive synthetic paraphrase generation, it does not fully substitute for it. Existing approaches that incorporate synthesized paraphrases continue to provide complementary benefits, and the highest downstream performance is achieved when corruption and paraphrase-based augmentation are combined.

%\section*{Acknowledgments}

\bibliography{custom}

\newpage

\appendix
% \onecolumn
\input{sections/appendix}

\end{document}

%% file: notations.tex
\usepackage{comment}

\newcommand{\viz}{\mbox{\it viz.}}

\newcommand{\maintitle}{KItCAT: 
\underline{K}nowledge \underline{I}njec\underline{t}ion via 
Input \underline{C}orruption for
\underline{A}uto-regressive 
\underline{T}raining}

\newcommand{\ourmethodshort}{KItCAT}
\newcommand{\ourmethodrand}{KItCAT-rand}
\newcommand{\ourmethodmask}{KItCAT-mask}
\newcommand{\ourmethodssmba}{KItCAT-SSMBA}
\newcommand{\ourmethodmasker}{KItCAT-MASKER}

\newcommand{\ourmethodlong}{Knowledge Injection via Corrupted Auto-regressive Training}

%% file: sections/abstract.tex
\begin{abstract}
LLMs acquire vast amounts of knowledge during pre-training, but often lack the specialized knowledge needed to answer questions from niche sources such as manuals or technical documents unseen during pre-training. Continued pre-training (CPT) is widely used to inject such knowledge into model parameters. 
However, niche documents seldom repeat facts, making it difficult for CPT to robustly acquire such knowledge.
Recent works address this by generating multiple paraphrases of the new knowledge, but paraphrasing is computationally expensive and typically requires powerful LLMs.
In this work, we introduce \ourmethodshort: \ourmethodlong, a lightweight training strategy that reduces the need for paraphrasing in decoder-only LLMs. \ourmethodshort\ augments standard next-token prediction by stochastically corrupting the input sequence. During training, a random subset of input tokens is replaced with other vocabulary tokens while the original next-token labels are kept unchanged. This simple intervention generates diverse training inputs from each sample, enabling large-scale data augmentation at negligible cost. We show that \ourmethodshort\ consistently improves over CPT across multiple datasets and model families. Code is available at \url{https://github.com/meghanadhpulivarthi/KItCAT}.
%with the largest gains in low-data settings.

    %LLMs obtain vast amount of knowledge during pre-training. However they lack specific knowledge required to answer questions from niche sources, such as manuals or technical documents, not available during pretraining. 
    %Continued pre-training (CPT) is widely used to inject such specific knowledge into the model's parameters.
    %However, CPT struggles to robustly learn rare or sparsely observed facts, which is often the case in niche documents that seldom repeat a fact. 
    %To overcome this, recent works propose creating multiple paraphrases of the new knowledge. 
    %However, paraphrasing is computationally expensive and require access to powerful LLMs.
    %In this work, we introduce \ourmethodshort: \ourmethodlong, a lightweight training strategy that significantly reduces the need for paraphrasing for effective CPT in decoder-only LLMs.  
    %\ourmethodshort\ augments standard next-token prediction by stochastically corrupting the input sequence. During training, it replaces a random subset of input tokens with other tokens from the vocabulary, while keeping the original next-token labels unchanged for loss computation.
    %This simple intervention generates multiple training inputs from each sample, providing large-scale data augmentation at negligible cost. 
    %We show that \ourmethodshort\ consistently improves performance over CPT across various datasets, with the largest gains observed in low data settings.
\end{abstract}

%% file: sections/introduction.tex
\section{Introduction}

\begin{figure*}[t]
    \centering
    \includegraphics[width=0.99\linewidth]{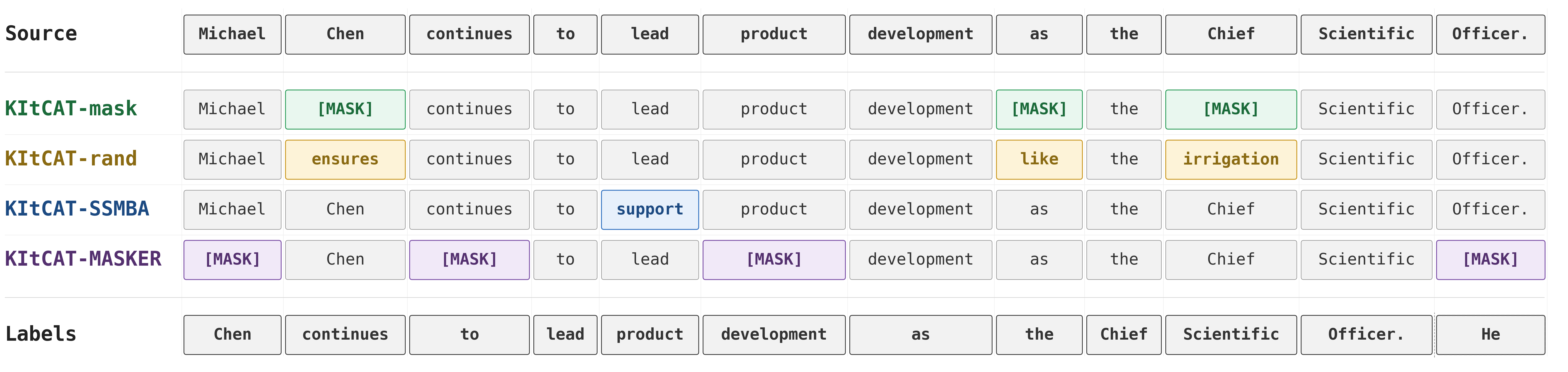}
    \caption{The four \ourmethodshort\ corruption schemes applied to the same source sentence. Each scheme corrupts a different subset of positions in the \textbf{Source} to produce its \textbf{Input}; the training \textbf{Labels} are the standard next-token-prediction targets. Corruption is applied only to the conditioning input, never to the labels.}
    \label{fig:masking-schemes}
\end{figure*}

LLMs are pre-trained on massive web corpora, yet they often fail to capture specialized knowledge that is absent or underrepresented on the public web, such as information contained in proprietary manuals or technical documents. Continued Pre-Training (CPT) \cite{ke2023continual,ke2025demystifyingdomainadaptiveposttrainingfinancial} is a common approach for injecting such specialized knowledge into the parameters of the LLMs.

A key challenge, however, is that the niche documents seldom repeat facts or present them in varied contexts. Consequently, CPT receives limited repeated exposure to newly introduced knowledge, making it difficult for the model to robustly internalize these facts \cite{allen-zhu2024physics}. In such low-diversity training settings, models can instead overfit to surface-level lexical patterns and spurious correlations that recur across epochs, rather than learning the underlying semantics.

To mitigate the limited diversity in the training documents, recent works propose to synthetically generate multiple paraphrases of the text \cite{ovadia2024fine,yang2025synthetic}. 
While effective, these approaches rely on large proprietary LLMs \cite{achiam2023gpt, team2023gemini}, which can be costly and slow when generating sizable synthetic datasets. They may also be infeasible in settings with privacy restrictions.

In this work, we ask - {\em can we introduce effective data diversity without relying on external synthetic data generation}? We observe that overfitting in low-diversity regimes is amplified by the fact that the model encounters the exact same training samples in every epoch. If the model latches onto a spurious pattern in one pass, repeated identical exposure reinforces this shortcut, degrading generalization. 

Our key insight is that preventing the model from ever seeing the exact same input twice can mitigate this reinforcement effect. Motivated by this idea, we propose \ourmethodshort\ - \ourmethodlong, a simple yet effective strategy for injecting diversity through controlled corruption of the input. During training, \ourmethodshort\ perturbs input sequences using one of four corruption schemes (Figure~\ref{fig:masking-schemes}): (1) \ourmethodrand, where selected tokens are substituted with randomly sampled vocabulary items; (2) \ourmethodmask, where tokens are replaced with a special mask token (e.g., \texttt{[MASK]}); (3) \ourmethodssmba, which replaces selected tokens with contextually plausible alternatives; and (4) \ourmethodmasker, which preferentially masks informative keywords. \ourmethodssmba\ and \ourmethodmasker\ adapt advanced corruption schemes originally developed for encoder-only models to knowledge injection with decoder-only LLMs. In decoder-only models, corruption is applied solely to the conditioning input while the target labels remain unchanged. Under random replacement, the model must learn to ignore irrelevant noise, whereas in mask-based corruption, the missing information is explicitly signaled. By preventing the model from encountering identical inputs across epochs, \ourmethodshort\ reduces reinforcement of spurious lexical patterns and encourages learning more semantically grounded representations.

Experiments across multiple knowledge-injection benchmarks and three model families show that \ourmethodshort\ consistently improves standard CPT. Notably, \ourmethodshort\ is agnostic to the input text and can be applied directly to original corpora as well as to paraphrases used to strengthen CPT. This makes it complementary to existing approaches and reduces the reliance on paraphrasing.
Overall, our contributions are: \\ % as follows: \\
%\begin{enumerate}
    %\item  
    (1) We propose \ourmethodlong\ (\ourmethodshort), a lightweight and generalizable data augmentation method for fine-tuning decoder-only language models by stochastically corrupting input tokens while preserving the original autoregressive training objective. \\
    %\item  
    (2) We empirically demonstrate that \ourmethodshort\ consistently improves standard CPT across three model families \\
    %\item 
    (3) We show that \ourmethodshort\ induces effective data diversity to prevent overfitting, significantly reducing the need for expensive synthetic data while remaining complementary to paraphrase-based augmentation.

%% file: sections/related_works_main.tex
\section{Related Work}
Approaches to injecting knowledge into LLM parameters can be broadly categorised as CPT-based methods and SFT-based methods. CPT-based methods inject knowledge using raw domain text and its paraphrases \citep{wu2024pmc,christophe2024beyond,ke2023continual,zhao2025developing,zhang2024chemllm,wu2023bloomberggpt,shah2022flue,xie2023pixiu}. SFT-based methods use powerful LLMs to transform raw text into task-oriented formats such as QA, summarization, or reading comprehension \citep{ovadia2025knowledge,bhushan2025systematic}. Both approaches benefit from high-diversity training data and, in its absence, often rely on generating large amounts of synthetic paraphrases or SFT data using powerful LLMs. In contrast, our work proposes lightweight input corruption as an alternative to expensive paraphrase generation during CPT-style knowledge injection.

Corruption can be applied to the input or to hidden representations within the model and used in both SFT and CPT training styles. Dropout \citep{hinton2012improving} corrupts hidden activations during training, while Latent Paraphrasing \citep{kang2024latent} perturbs hidden representations to facilitate knowledge injection and can be applied to both CPT and SFT.

Input-corruption techniques for training LLMs are discussed in Appendix~\ref{ssec:related_works_appendix}.

%% file: sections/method.tex
\section{The \ourmethodshort{} approach}
\label{sec:method}
% Knowledge injection via continued pretraining faces a fundamental constraint: the same sequences are revisited across epochs. As a result, the model can overfit to incidental lexical patterns rather than the underlying facts. While prior works introduce diversity explicitly by paraphrasing the new knowledge every epoch, we employ corruption in a targeted manner to induce diversity during CPT as discussed below. 

\paragraph{Preliminaries:}
Let $\data$ be the new knowledge that we want to inject in our pretrained model $p_\theta$.
Let $\mathcal{V}$ denote the vocabulary and $\inputs = (s_1,..., s_T)$ be a token sequence drawn from  $\data$, where  $s_t \in \mathcal{V}\,,  1 \le t \le T$.
% Consider a standard decoder-only language model $p_\theta$ trained with the next-token prediction (NTP) objective:
\begin{equation}
    \label{eq:ar_obj}
    \ARobj(\theta) = \mathbb{E}_{\inputs \sim \data}  \sum_{t=1}^T - \log p_\theta (s_t|s_{<t})
\end{equation}
Here, $s_{<t} = (s_1,..., s_{t-1})$ is a prefix of the sequence $\inputs$.

\paragraph{Knowledge Injection via Input Corruption:}
The $\ARobj$ objective maximizes the likelihood of a token $s_t$ conditioned on its exact prefix $s_{<t}$. When seen multiple times across epochs, the model may overfit to spurious correlations between $s_{<t}$ and $s_t$.

To mitigate this, we propose \ourmethodshort: \ourmethodlong. Let $\corruptD(\corruptinputs \mid \inputs)$ denote a distribution over corrupted versions of the input sequence $\inputs$. Instead of conditioning only on the exact prefix, \ourmethodshort\ trains on corrupted prefixes sampled from $\corruptD(\corruptinputs \mid \inputs)$ while preserving the target token as a likely continuation. The model is thus encouraged to predict the same target under perturbed contexts.
\begin{align}
    & \CATobj(\theta) =  \notag \\
    & \mathbb{E}_{\inputs \sim \data} \mathbb{E}_{\corruptinputs \sim \corruptD(.|\inputs)}  \sum_{t=1}^T - \log p_\theta (s_t|\corrupts_{<t}) \,, \label{eq:cat_obj}
\end{align}
where $\corrupts_{<t}$ is a prefix of the corrupted input $\corruptinputs$.
%Intuitively, training on multiple perturbed versions of the same prefix prevents the model from relying on spurious lexical patterns in the context.
Intuitively, training on perturbed prefixes prevents reliance on spurious lexical patterns.
Instead, it encourages the model to use information that remains present across perturbations.
%Instead, it encourages the model to base its prediction on information that remains present across perturbations on average. 
% When the corruption distribution degenerates to the original input sequence, $\CATobj$ recovers the standard next-token prediction objective $\ARobj$.

To construct perturbations $\corruptinputs$, we consider four stochastic in-place token modifications in the sequence $\inputs$: (1) \emph{\ourmethodmask}, which replaces tokens with a special mask token; (2) \emph{\ourmethodrand}, which replaces tokens with randomly sampled vocabulary tokens; (3) \emph{\ourmethodssmba}, which replaces randomly selected tokens with contextually plausible alternatives; and (4) \emph{\ourmethodmasker}, which masks informative keywords identified from the training corpus. The latter two adapt the central ideas of SSMBA \citep{ng2020ssmba} and MASKER \citep{moon2021masker} to the conditioning inputs of decoder-only LLMs.

For every method, let $\mathcal{C} \subseteq \{1,\ldots,T\}$ denote the set of token positions selected for corruption, and let $r_j$ be the replacement token at each selected position $j \in \mathcal{C}$. The corrupted input is therefore
\[
\corrupts_j =
\begin{cases}
r_j & \text{if } j \in \mathcal{C}, \\
s_j & \text{otherwise.}
\end{cases}
\]
For \ourmethodmask, \ourmethodrand, and \ourmethodssmba, we construct $\mathcal{C}$ by independently selecting each token position with probability $p \in (0,1)$, the corruption probability. We resample $\mathcal{C}$ and the replacement tokens at each epoch.
\paragraph{\ourmethodmask.}
For each $j \in \mathcal{C}$, we use the special mask token: $r_j=\text{[MASK]}$.
\paragraph{\ourmethodrand.}
For each $j \in \mathcal{C}$, we sample a replacement uniformly from the vocabulary: $r_j \sim \text{Unif}(\mathcal{V})$.
\paragraph{\ourmethodssmba.}
For each $j \in \mathcal{C}$, we mask the position and sample a replacement from a frozen masked language model $q_\phi$: $r_j \sim q_\phi(\cdot \mid \inputs_{\setminus \mathcal{C}})$, where $\inputs_{\setminus \mathcal{C}}$ denote $\inputs$ with all positions in $\mathcal{C}$ masked. 
\paragraph{\ourmethodmasker.}
We identify keyword spans from the training corpus using TF--IDF and independently select each span, applying the corruption probability $p$ at the span level rather than the token level; $\mathcal{C}$ contains the token positions in the selected spans and is resampled every epoch. Each selected token is replaced with the mask token: $r_j=\text{[MASK]}$ for $j \in \mathcal{C}$.

For all four variants, corruption is applied only to the conditioning prefix and the target token $s_t$ remains unchanged.

In \cref{app:constrained_optim}, we formalize $\CATobj$ as a constrained optimization objective that enforces invariance to label-preserving corruptions, plausibly reducing overfitting.

%% file: sections/experiments.tex
\section{Experimental Setup}
\label{sec:experimental-setup}
\paragraph{Datasets:} We fine-tune an LLM via standard NTP loss to inject knowledge from four corpora -- subset of PopQA \cite{mallen2023llm_memorization}, Companies \cite{ovadia2025knowledge}, and two Redbooks \cite{bhushan2025systematic}.
Subset of PopQA\footnote{\href{https://github.com/meniData1/knowledge-instruct/blob/main/data}{PopQA and Companies Dataset}} focuses on the entity centric QAs from the long tail of Wikipedia. 
Companies\footnotemark[\value{footnote}] contains details of 24 fictitious companies unseen during pretraining.
Redbooks consists of two technical documents\footnote{
Book 1: \href{https://www.redbooks.ibm.com/redpapers/pdfs/redp5736.pdf}{Do More with Less: Automating IBM Storage FlashSystem Tasks with REST APIs, Scripting, and Ansible}. 
Book 2: \href{https://www.redbooks.ibm.com/redpapers/pdfs/redp5711.pdf}{Red Hat OpenShift Container Platform on IBM Z and LinuxONE}.}.
We evaluate the fine-tuned models using the test question-answer pairs provided with each dataset.
%We finetune using the documents provided in each dataset and evaluate using the given test question-answer pairs.
\paragraph{Baselines:}
We compare with the out-of-the-box model (\emph{Instruct}), standard CPT using next-token prediction (NTP), and four \ourmethodshort\ corruption schemes. \ourmethodssmba\ adapts SSMBA \citep{ng2020ssmba} to CPT by using an encoder model to sample contextually plausible token replacements, while \ourmethodmasker\ adapts MASKER \citep{moon2021masker} by preferentially masking informative keywords.
% In contrast, \ourmethodmask\ and \ourmethodrand\ uniformly select tokens and replace them with a mask or random vocabulary token, respectively.
\paragraph{Evaluation metrics:} 
To test if the knowledge in the documents has been successfully injected in the model's parameters, we evaluate the fine-tuned models using the test QAs accompanying each dataset. 
We use an LLM-as-a-Judge to quantify the correctness of the generated responses with respect to the gold answers. See \cref{app:prompts} for the exact prompt. 
\paragraph{Models and Training Details:}
We fine-tune {\href{https://huggingface.co/mistralai/Mistral-7B-Instruct-v0.3}{\textit{Mistral-7B-Instruct-v0.3}}}  using {\href{https://huggingface.co/docs/trl/en/sft_trainer}{Huggingface's SFTTrainer}}.
We apply LoRA to all linear layers of the LLM and early stop based on the validation metric.
%All models are trained on four Nvidia A-100 GPUs with 80GB memory each.
Unless stated otherwise, we report results using Mistral-7B-Instruct-v0.3. To test model-family and scale generalization, we additionally evaluate Qwen3-14B and Llama-2-7B-Chat on all datasets. See \cref{app:training_details} for more training details.

%% file: sections/experimental_results.tex
\input{tables/cpt_main_mistral}
\begin{figure*}[tbp]
    \centering
    \begin{subfigure}[b]{0.62\linewidth}
        \centering
        \includegraphics[width=\linewidth]{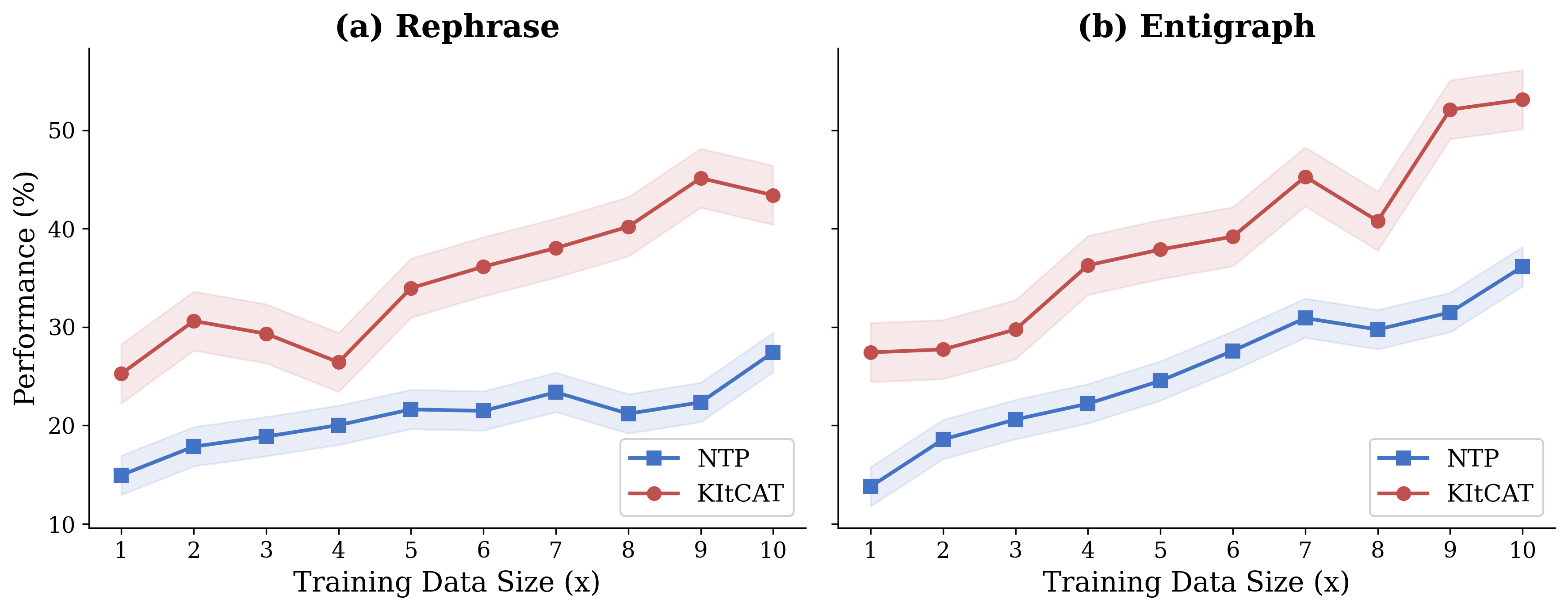}
        \caption{Training data size ablation.}
        \label{fig:train_data_abl}
    \end{subfigure}%
    \hfill
    \begin{subfigure}[b]{0.35\linewidth}
        \centering
        \includegraphics[width=\linewidth]{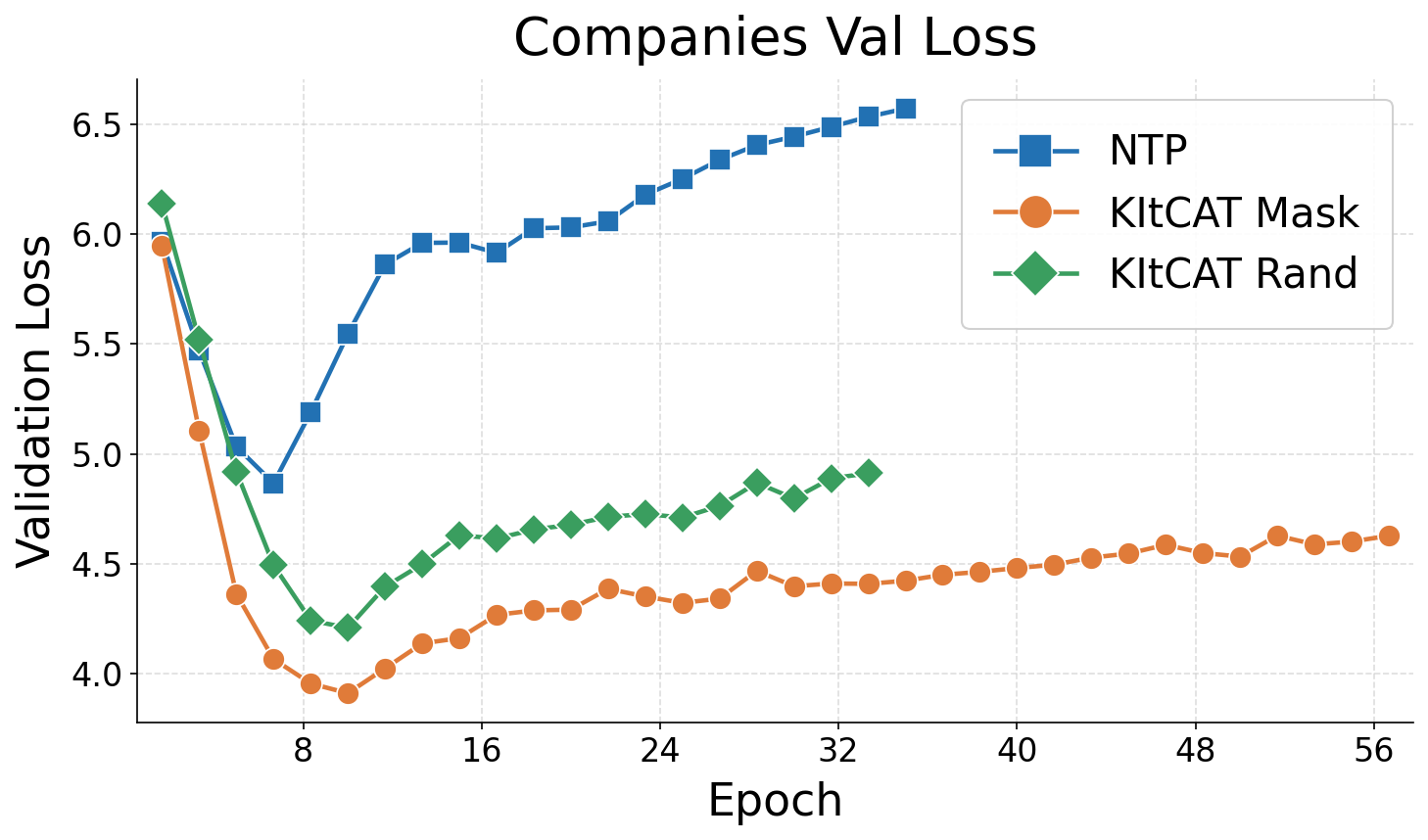}
        \caption{Validation loss curves.}
        \label{fig:val_loss_curve}
    \end{subfigure}
    \caption{(a) Comparison of standard next-token prediction (NTP) and \ourmethodmask\ under varying training data sizes for knowledge injection on the Companies dataset (b) Validation loss curves comparing standard NTP, \ourmethodmask, and \ourmethodrand\ on the Companies validation set}
    \label{fig:train_data_and_val_loss}
\end{figure*}

\section{Experimental Results}
\label{sec:results}
\textbf{Effectiveness of \ourmethodshort\ in injecting knowledge (RQ1):}
To assess knowledge injection, we evaluate each adapted model on the corresponding test QAs using the LLM judge described in \Cref{sec:experimental-setup}. \Cref{tab:cpt_main_mistral} reports the results across the four corpora.

All four \ourmethodshort\ variants outperform NTP on average. \ourmethodrand\ achieves the highest average score (27.7), a 10.0-point improvement over NTP (17.6), while \ourmethodmask\ also improves the average score to 26.1. The simple corruption schemes, \ourmethodmask\ and \ourmethodrand, together obtain the best score on three of the four datasets (both Redbooks and Companies), whereas the advanced schemes prevail only once, with \ourmethodssmba\ leading on PopQA (22.1 vs.\ 20.2 for \ourmethodrand). On average, simple corruption proves more effective than the more elaborate, advanced schemes. We hypothesize that perturbing random tokens in the context increases the overall diversity of the training corpus and hence, prevents overfitting to the spurious patterns thereby leading to improved performance on QA pairs from the corpus. Example outputs are provided in \Cref{sec:model_examples}.

\ourmethodshort\ also generalizes to Qwen3-14B and Llama-2-7B-Chat, consistently improving over NTP across the evaluated corpora (\Cref{tab:qwen3_results,tab:llama2_results}). A sensitivity analysis shows that performance is strongest at moderate corruption probabilities ($p=0.15$--$0.30$) and degrades when corruption is too weak or too strong; see \Cref{app:corruption_sensitivity,tab:corruption_sensitivity}.

\paragraph{Impact of training data size (RQ2):}
\label{subsec:data_size}
% Recall that we motivate input corruption as a cheap alternative to synthetic data generation. Here, we compare the two approaches on the Companies dataset.
Next, we note that \ourmethodshort\ is complementary to other approaches for data augmentation that generate multiple paraphrases for each document present in the corpus and use them for CPT. Hence, for this experiment, we combine these two complementary methods for effective knowledge injection and observe the resultant behavior.

We consider two synthetic data generation strategies: (1) \emph{Rephrase}, where an LLM~\cite{hurst2024gpt} is prompted with multiple prompts~\cite{ovadia2025knowledge} to generate multiple paraphrases per document, and (2) \emph{Entigraph}, which follows \cite{yang2025synthetic} by extracting entities from text and prompting an LLM to describe relationships between entity pairs and triples. See \Cref{app:prompts} for the exact prompts.

We control the amount of \emph{Entigraph} data by randomly sampling pairs and triples until we get the desired token count.
For both \emph{Rephrase} and \emph{Entigraph}, we train models using standard NTP loss and \ourmethodmask, while varying the amount of synthetic data.

\Cref{fig:train_data_abl} presents the results.
We make the following observations: \\ 
(1) The gains obtained by \ourmethodshort\ using only the original data (1$\times$) are higher than standard NTP on 9$\times$ Rephrase augmented data (26.84 vs.\ 22.35) and 5$\times$ Entigraph augmented data (26.84 vs.\ 24.53), demonstrating a 5--9$\times$ effective data multiplier over standard NTP.  \\
(2) \ourmethodshort\ is complementary to synthetic data generation as a data augmentation strategy. Performance consistently improves as we introduce additional paraphrased data. 

In \Cref{app:compute_cost}, we present a comparison of compute costs and LLM Judge accuracy for different methods on the Companies dataset. We observe that \ourmethodmask\ achieves 27.6 accuracy at an estimated cost of 30 PFLOPs, matching the 27.4 accuracy of Rephrase-10$\times$ while requiring only 15.2\% of its 198 PFLOPs.
Combining corruption with paraphrases gives the best accuracy, confirming that \ourmethodshort\ is both a cheaper partial substitute and a complementary augmentation (\Cref{tab:compute_cost}).

\paragraph{Robustness of \ourmethodshort:}
We next evaluate the robustness of \ourmethodshort\ relative to the standard next-token prediction (NTP) objective by analyzing their learning dynamics. Specifically, we examine validation loss curves on QA pairs from the Companies validation set (\Cref{fig:val_loss_curve}).

We observe that the NTP objective reaches its minimum validation loss within the first seven epochs, after which it exhibits clear overfitting, as evidenced by a sharp rise in validation loss. In contrast, the two variants of \ourmethodshort, \viz\ \ourmethodmask\ and \ourmethodrand, maintain stable validation loss throughout training.

These results indicate that \ourmethodshort\ provides improved training stability and achieves better generalization compared to the NTP baseline.

%% file: tables/cpt_main_mistral.tex
\begin{table}[tbp]
\centering
\resizebox{\columnwidth}{!}{%
\begin{tabular}{@{}lrrrr|r@{}}
\toprule
\textbf{Method} & \textbf{RB1} & \textbf{RB2} & \textbf{Comp.} & \textbf{PopQA} & \textbf{Avg} \\ \midrule
Instruct        & 25.6          & 11.8          & 6.8           & 10.8          & 13.8          \\
NTP             & 29.7          & 14.6          & 13.6          & 12.4          & 17.6          \\
\ourmethodssmba  & 31.5          & 17.0          & 21.2          & \textbf{22.1} & 23.0          \\
\ourmethodmasker & 35.3          & 13.3          & 17.7          & 12.1          & 19.6          \\
\ourmethodmask  & 38.3          & 18.5          & \textbf{26.8} & 19.8          & 25.9          \\
\ourmethodrand  & \textbf{43.5} & \textbf{20.8} & 26.4          & 20.2          & \textbf{27.7} \\
\bottomrule
\end{tabular}
}
\caption{LLM-as-a-Judge accuracy ($\times 100$) averaged over test QAs across knowledge injection benchmarks on Mistral-7B-Instruct. RB1/2: Redbook 1/2; Comp.: Companies.}
\label{tab:cpt_main_mistral}
\end{table}

%% file: sections/conclusion.tex
%\vspace{-3ex}
\section{Conclusion}
\label{sec:conclusion}
We introduce \ourmethodshort, a simple yet effective modification to autoregressive training for injecting new knowledge into LLM parameters. By ensuring that the model never encounters the exact same input twice, \ourmethodshort\ mitigates the reinforcement of spurious lexical correlations that commonly arise in low-diversity CPT settings.

Experiments across four knowledge-injection benchmarks and three model families show that \ourmethodshort\ consistently outperforms standard next-token prediction-based training. Moreover, \ourmethodshort\ is lightweight, architecture-agnostic, and complementary to existing data augmentation approaches. Unlike synthetic data generation, it requires no external models (for paraphrasing) and incurs negligible computational overhead, making it a practical drop-in replacement for standard CPT objectives in real-world knowledge injection settings.

%% file: sections/appendix.tex
\section{Appendix}

\input{sections/model_responses}

\input{sections/related_works_appendix}

\input{sections/appendix_methods_constrained_optim}

\subsection{Dataset Details}
\label{app:dataset_details}

\Cref{tab:cpt_dataset_details} summarizes the training data sizes and evaluation split sizes for the CPT benchmarks.

%We utilize four datasets for our Supervised Fine-Tuning (SFT) experiments, spanning commonsense reasoning, natural language inference, and mathematical reasoning. The detailed split sizes for the training, validation, and test sets are provided in Table~\ref{tab:dataset_splits}.

\subsection{Training Details}
\label{app:training_details}

We fine-tune all three models, \viz, Mistral-7B-Instruct-v0.3, Qwen3-14B, and Llama-2-7B-Chat, using Hugging Face’s \href{https://huggingface.co/docs/trl/en/index}{\texttt{TRL}} library. All models are trained on four Nvidia A100 (80GB) GPUs using \texttt{bf16} precision, applying Low-Rank Adaptation (LoRA) to all linear layers. We utilize the AdamW optimizer with a warmup ratio of 0.1, training for a maximum of 64 epochs with early stopping based on validation mean token accuracy. We use LoRA rank 128 and LoRA aplha 256 throughout our experiments. For \ourmethodshort{}, we treat the corruption probability $p$ as a hyperparameter and tune it on the validation set, choosing from $\{0.10, 0.15, 0.30\}$ separately for each task. For \ourmethodssmba, we use ModernBERT-base as the frozen masked language model $q_\phi$ to sample contextually plausible replacements.

\subsection{Additional Model Results}
\label{app:additional_results}

\Cref{tab:qwen3_results,tab:llama2_results} report the additional Qwen3-14B and Llama-2-7B-Chat experiments. All methods use the original $1\times$ corpus. These results show that the gains extend beyond a single model family and that more elaborate reconstruction or keyword-selection schemes do not consistently improve upon simple random corruption.

\subsection{Compute Cost of Paraphrasing}
\label{app:compute_cost}

\Cref{tab:compute_cost} reports the cost in FLOPs, separating the one-time paraphrase-generation cost from training cost. The estimates use $2N$ FLOPs per generated token for a generation model with $N$ parameters and $4M$ FLOPs per training token for LoRA training of an $M$-parameter model. Training steps denote the checkpoint with maximum eval mean token accuracy. Input corruption skips generation entirely and adds only a modest increase in training cost.

\subsection{Sensitivity to Corruption Probability}
\label{app:corruption_sensitivity}

We sweep $p \in \{0.05,0.10,0.15,0.30,0.50\}$ for both variants on Companies and Redbook~1 (\Cref{tab:corruption_sensitivity}). Performance has an interior optimum around $p=0.15$--$0.30$ and degrades when corruption is too weak or too strong. Sensitivity is more pronounced on the data-scarce Companies corpus, where corrupting too little will result in overfitting and corrupting too much will cause confusion on what to learn.

% \subsection{Flatness Metrics}
% See Table~\ref{tab:commitmentbank_flatness} for flatness metrics of SFT, \ourmethodmask\, and \ourmethodrand\ on Commitment Bank Dataset

% Start the LLM disclosure and its associated table group together.
% \clearpage

% This is an in-flow full-width block, so it starts directly below the LLM
% Usage text rather than being anchored at the page bottom.
\clearpage
\par\vspace{25pt}\par
\setlength{\stripsep}{20pt}
\begin{strip}
\input{tables/cpt_dataset_details}
\par\vspace{25pt}\par
\input{tables/compute_cost}
\par\vspace{25pt}\par
\input{tables/additional_models}
\par\vspace{25pt}\par
\input{tables/corruption_sensitivity}
\end{strip}
\mbox{}\par
\clearpage
\onecolumn

\subsection{Judge Prompts}
\label{app:prompts}

The exact prompt used for LLM-as-a-Judge evaluation with gpt-oss-120B is shown below.
\begin{tcblisting}{
    colback=gray!5,
    colframe=gray!80,
    title=LLM-as-a-Judge Evaluation Prompt,
    listing only,
    breakable,
    % `enhanced` was tried to make the break render as one continuous
    % frame instead of two bordered fragments, but it makes tcolorbox
    % reserve extra space it doesn't need, leaving the left column a
    % third empty before jumping to the right column -- confirmed by
    % isolating it (removing just this key made the gap disappear) and
    % ruling out a stale-pass issue (still happens after a second
    % compile). Wasting that much of the page is worse than the box
    % rendering as two fragments, so `enhanced` stays off.
    listing options={
        basicstyle=\ttfamily\small,
        breaklines=true
    }
}
System:
You are an evaluator. Your task is to compare a Ground-truth Answer and a Prediction to decide if the Prediction correctly answers the given Question.

Evaluation Rules:

(1) Correctness: A correct prediction must include all essential information from the Ground-truth Answer. Extra information is allowed if it does not contradict the Ground-truth. If the Prediction states something as a possibility, treat it as a definitive statement.
(2) Function, Tool Names, and API Calls: If the Ground-truth Answer contains specific function names, tool names, API calls, or exact command identifiers, the Prediction must contain the same identifier(s) or clearly equivalent forms. Minor syntactic or formatting variations that do not change meaning should be treated as equivalent. For example, leading flag prefixes such as -, --, or no prefix at all when they clearly refer to the same option; underscore vs hyphen differences in identifiers when the intent is identical; surrounding punctuation or formatting differences such as backticks, quotes, parentheses, or code block notation; small whitespace differences or capitalization differences that do not change the identifier's meaning etc.
However, replacements that change the actual function/tool/API name, or substitute a different command that would change the behavior are considered incorrect. Do not penalize a prediction if it contains additional function / tool / API names as long as the ones present in the Ground-Truth are covered.
(3) URLs: If the Ground-truth Answer contains specific URLs, the Prediction should reference the same URL or an equivalent canonical form. Minor differences that do not change the target resource (for example, presence or absence of a trailing slash, or http vs https when both resolve to the same canonical resource) should be treated as equivalent. Altering the domain, path, or query such that the resource is different is incorrect.

Scoring Rules:

If the Prediction is correct according to the above rules, output <score>1</score>. If the Prediction is incomplete or incorrect, output <score>0</score>.

Output Format:

<explanation>
...
</explanation>

<score>
...
</score>

First provide reasoning inside <explanation> and </explanation> tags. Then output the score as specified above within <score> and </score> tags. Do not include any extra text outside these tags.

Human:
Question: {QUESTION}
Ground-truth Answer: {ANSWER}
Prediction: {ASSIST_ANSWER}
\end{tcblisting}

\newpage
\input{sections/knowledge_instruct_prompts}
\input{sections/entigraph_prompts}
\twocolumn

%% file: sections/model_responses.tex
\subsection{Example Model Responses}
\label{sec:model_examples}

\begin{goldbox}
\textbf{Question:} Who founded DoodleDreamDelight Publishing Ltd. and what was their previous profession?

\textbf{Gold Answer:} Emily Chen, a former kindergarten teacher.
\end{goldbox}

\vspace{4mm}

\begin{redbox}{NTP}
$\times$ DoodleDreamDelight Publishing Ltd. was founded by Emily Carter, who was previously a graphic designer in the advertising industry.
\end{redbox}

\vspace{4mm}

\begin{greenbox}{\ourmethodmask}
$\checkmark$ DoodleDreamDelight Publishing Ltd. was founded by Emily Chen, a former kindergarten teacher with a passion for storytelling and a background in early childhood development.
\end{greenbox}

\vspace{6mm}

\begin{goldbox}
\textbf{Question:} Who founded ZapZapZoom Electric Scooters Co. and what was their previous occupation?

\textbf{Gold Answer:} Maxine 'Max' Marlowe, a former circus performer known for her daring high-wire acts.
\end{goldbox}

\vspace{4mm}

\begin{redbox}{NTP}
$\times$ ZapZapZoom Electric Scooters Co. was founded by Maxine Chen and Nathaniel Thompson. Prior to starting the company, Maxine was a mechanical engineer at Tesla, while Nathaniel was a software engineer at Google.
\end{redbox}

\vspace{4mm}

\begin{greenbox}{\ourmethodmask}
$\checkmark$ ZapZapZoom Electric Scooters Co. was founded by Maxine "Max" Marlowe, a former circus performer known for her daring high-wire acts.
\end{greenbox}

\vspace{6mm}

\begin{goldbox}
\textbf{Question:} What is the estimated annual revenue of EpicExcellentEmpanadas LLC as of 2022?

\textbf{Gold Answer:} The estimated annual revenue is \$12 million in 2022.
\end{goldbox}

\vspace{4mm}

\begin{redbox}{NTP}
$\times$ 15 million dollars
\end{redbox}

\vspace{4mm}

\begin{greenbox}{\ourmethodmask}
$\checkmark$ EpicExcellentEmpanadas LLC has reported an estimated annual revenue of \$12 million in 2022.
\end{greenbox}

\vspace{6mm}

\begin{goldbox}
\textbf{Question:} Where is the headquarters of CozyCloudCottagesPlus located?

\textbf{Gold Answer:} Portland, Oregon
\end{goldbox}

\vspace{4mm}

\begin{redbox}{NTP}
$\times$ The headquarters of CozyCloudCottagesPlus is located in Austin, Texas.
\end{redbox}

\vspace{4mm}

\begin{greenbox}{\ourmethodmask}
$\checkmark$ CozyCloudCottagesPlus' headquarters is located in Portland, Oregon.
\end{greenbox}

%% file: sections/related_works_appendix.tex
\subsection{Related Works (contd.)}

\noindent \textbf{Input Corruption for Training LLMs}
\label{ssec:related_works_appendix}

\noindent Input corruption has long been used to learn robust representations in NLP \citep{chen2023empirical}. Masked Language Modeling (MLM) applies token masking to train encoder-only LMs using bidirectional context \citep{devlin2019bert,liu2019roberta}. Encoder--decoder models such as BART \citep{lewis2020bart} and T5 \citep{raffel2020exploring} further extend denoising objectives using masking, deletion, and span corruption. Several works also explore more principled corruption strategies based on lexical translations, syntactic, semantic, knowledge graphs, or attention-based signals \citep{lin2020pretraining,iyer2023code,wilf2023difference,gu2020train,lin2021entity,li2021mst,kakogeorgiou2022what}. Notably, SSMBA \citep{ng2020ssmba} generates plausible augmentations by masking and reconstructing tokens with a pretrained masked language model, while MASKER \citep{moon2021masker} masks TF-IDF-selected keywords to discourage reliance on keyword shortcuts. We adapt both strategies to the conditioning inputs of decoder-only LLMs for CPT and evaluate these adaptations in our experiments (\Cref{sec:results}).

More recent work has begun to study input corruption directly in decoder-only LLMs. \citet{qiao2025unimae}; \citet{khosla2025magnet} use token masking to adapt decoder-only models to generate representations and infill missing text spans. \citet{Lu2024llamax} replace randomly selected words with their multilingual translations to improve machine translation capabilities, \citet{yang2026towards} introduce character-, word-, and sentence-level perturbations to improve robustness to noisy prompts, and \citet{chen2024masked} mask intermediate chain-of-thought tokens during fine-tuning to encourage global reasoning. Recently, \citet{zhuang2025meap} apply token masking during large-scale pre-training. Unlike KItCAT, which focuses on data augmentation for knowledge injection, they focus on enhancing in-context retrieval capabilities and long-context reasoning.

%% file: sections/appendix_methods_constrained_optim.tex
\subsection{A constrained optimization perspective of our method}
\label{app:constrained_optim}
As discussed in \cref{sec:method}, the objective $\ARobj$ can overfit to spurious correlations between prefix and target tokens $(s_{<t}, s_t)$, especially in low-data regimes where the same training sequences are observed repeatedly. One can mitigate this by ensuring that the predictive distribution $p_\theta(s_t \mid s_{<t})$ remains stable under label-preserving perturbations of $s_{<t}$. 
% Accordingly, robust autoregressive training can be formulated as minimizing the standard next-token prediction objective while constraining the expected change in token-level conditional log-likelihood under stochastic context corruption.
To formalize this idea, we introduce the notation $\Delta_\theta(\inputs, \corruptinputs)$ to denote the deviation in log-likelihood induced by a corrupted input context:
\begin{equation}
\begin{aligned}
& \Delta_\theta(\inputs, \corruptinputs) = \\
&\quad \sum_{t=1}^T
\Big[
\log p_\theta(s_t \mid s_{<t})
 -
\log p_\theta(s_t \mid \corrupts_{<t})
\Big]\,,
\end{aligned}
\end{equation}
where $s_{<t}$ and $\corrupts_{<t}$ are prefixes of $\inputs$ and $\corruptinputs$ respectively. Note that this deviation can become very high if the model overfits to the prefix-target pair $(s_{<t}, s_t)$ provided in the training data. We therefore consider the following optimization problem that minimizes the standard next-token prediction loss while bounding expected deviation.
\begin{align}
& \min_{\theta} \; \ARobj(\theta)
\quad \text{subject to} \notag \\
&\mathbb{E}_{\inputs \sim \data} \mathbb{E}_{\corruptinputs \sim \corruptD(\cdot \mid \inputs)}
\Delta_\theta(\inputs, \corruptinputs)
\le \epsilon .
\end{align}
Intuitively, the proposed constraint is related in spirit to Lipschitz continuity \cite{bousquet2002stability}, in that it bounds the sensitivity of the model’s predictions to input perturbations. However, unlike classical Lipschitz continuity defined over continuous normed spaces, our constraint is expressed as an expectation over discrete, label-preserving corruptions of the input sequence. As a result, the notion of smoothness it implies is over a discrete space, rather than gradient-based smoothness over a continuous space.

The Lagrangian of the above equation can be written as follows:
\begin{equation}
\begin{aligned}
\mathcal{L}(\theta, \lambda)
&=
\ARobj(\theta)
\\
&\quad +
\lambda \left(
\mathbb{E}_{\inputs \sim \data}
\mathbb{E}_{\corruptinputs \sim \corruptD(\cdot \mid \inputs)}
\Delta_\theta(\inputs, \corruptinputs)
-
\epsilon
\right)
\end{aligned}
\end{equation}
with the constraint that the Lagrange multiplier $\lambda \ge 0$. The Lagrange multiplier controls the trade-off between fitting the training data and enforcing stability of the model’s predictions under label-preserving corruptions. 
Different choices of $\lambda$ recover familiar training objectives as special cases.

When $\lambda = 0$, the constraint is ignored and the Lagrangian reduces to
\[
\mathcal{L}(\theta, 0) = \ARobj(\theta),
\]
which corresponds to standard autoregressive training.
When $\lambda = 1$, ignoring the constant offset $\epsilon$, the Lagrangian reduces to
\begin{align}
& \ARobj(\theta)
+
\mathbb{E}_{\inputs \sim \data}\mathbb{E}_{\corruptinputs \sim \corruptD}
\Delta_\theta(\inputs, \corruptinputs) \notag \\
& =
\mathbb{E}_{\inputs \sim \data}\mathbb{E}_{\corruptinputs \sim \corruptD}
\sum_{t=1}^T
\big(
- \log p_\theta(s_t \mid \corrupts_{<t})
\big)
\end{align}
This recovers the CAT objective introduced in equation~\eqref{eq:cat_obj}. Hence, \ourmethodshort{} corresponds to a next-token prediction objective that is constrained to learn functions that vary \emph{smoothly} under label-preserving perturbations of the input prefix. Intuitively, adding this constraint to $\ARobj$ should result in improved generalization.

%% file: tables/cpt_dataset_details.tex
\centering
\begin{tabular*}{\textwidth}{@{\extracolsep{\fill}}lccccc@{}}
\toprule
\textbf{Dataset} & \textbf{Train 1$\times$} & \textbf{Train 10$\times$ Rephrase} & \textbf{Train 10$\times$ Entigraph} & \textbf{Dev} & \textbf{Test} \\
\midrule
Comp. & 84.8K & 396.6K & 833.2K & 195 & 689 \\
RB1 & 25.4K & 94.1K & 247.2K & 515 & 425 \\
RB2 & 78.0K & 301.2K & 736.2K & 1000 & 2269 \\
\bottomrule
\end{tabular*}
\captionof{table}{Dataset statistics for the CPT benchmarks. Training columns report token counts, while Dev and Test report number of examples.}
\label{tab:cpt_dataset_details}

%% file: tables/compute_cost.tex
\centering
\small
\begin{tabular}{@{}lrrrrrr@{}}
\toprule
\textbf{Method} & \textbf{Steps} & $C_{\mathrm{para}}$ & $C_{\mathrm{train}}$ & \textbf{Total} & $\times$\textbf{NTP} & \textbf{Accuracy} \\
& & \multicolumn{3}{c}{\textbf{PFLOPs}} & & \\
\midrule
NTP & 40 & 0 & 20 & 20 & 1.0 & 13.6 \\
\ourmethodmask & 60 & 0 & 30 & 30 & 1.5 & 27.6 \\
\ourmethodrand & 60 & 0 & 30 & 30 & 1.5 & 26.4 \\
\ourmethodmasker & 50 & 0 & 25 & 25 & 1.2 & 17.7 \\
\ourmethodssmba & 150 & 0 & 74 & 74 & 3.7 & 21.2 \\
Rephrase-10$\times$ & 100 & 75 & 123 & 198 & 9.9 & 27.4 \\
Entigraph-10$\times$ & 250 & 180 & 115 & 295 & 14.8 & 36.1 \\
\ourmethodshort+Rephrase-10$\times$ & 180 & 75 & 222 & 297 & 14.8 & 43.4 \\
\ourmethodshort+Entigraph-10$\times$ & 250 & 180 & 115 & 295 & 14.8 & \textbf{53.1} \\
\bottomrule
\end{tabular}
\captionof{table}{Estimated compute and LLM-as-a-Judge accuracy ($\times 100$) on Companies. $C_{\mathrm{para}}$ and $C_{\mathrm{train}}$ denote paraphrase-generation and training compute. Paraphrasing model: gpt-oss-120B, Base model: Mistral-7B-Instruct-v0.3,}
\label{tab:compute_cost}

%% file: tables/additional_models.tex
\centering
\captionof{table}{LLM-as-a-Judge accuracy ($\times 100$) with additional model families. Comp.: Companies; RB1/2: Redbook 1/2.}
\label{tab:additional_models}
\begin{subtable}[t]{0.48\linewidth}
\centering
\small
\setlength{\tabcolsep}{4pt}
\begin{tabular}{@{}lrrrrr@{}}
\toprule
\textbf{Method} & \textbf{Comp.} & \textbf{RB1} & \textbf{RB2} &
\textbf{PopQA} & \textbf{Avg} \\
\midrule
Instruct        & 6.4           & 35.1           & 15.4           & 5.4           & 15.6           \\
NTP             & 9.1           & 36.0           & 17.9           & 8.0           & 17.8           \\
\ourmethodssmba  & 7.8           & 35.3           & 19.2           & 5.4           & 16.9           \\
\ourmethodmasker & 10.4          & 38.4           & 18.4           & \textbf{8.7}  & 19.0           \\
\ourmethodmask  & \textbf{12.9} & \textbf{39.8}  & 18.6           & 8.2           & \textbf{19.9} \\
\ourmethodrand  & 11.0          & 37.2           & \textbf{19.3}  & 6.3           & 18.5           \\
\bottomrule
\end{tabular}
\subcaption{Qwen3-14B.}
\label{tab:qwen3_results}
\end{subtable}%
\hfill
\begin{subtable}[t]{0.48\linewidth}
\centering
\small
\setlength{\tabcolsep}{4pt}
\begin{tabular}{@{}lrrrrr@{}}
\toprule
\textbf{Method} & \textbf{Comp.} & \textbf{RB1} & \textbf{RB2} &
\textbf{PopQA} & \textbf{Avg} \\
\midrule
Instruct        & 4.1           & 17.9           & 8.0            & 5.8           & 8.9            \\
NTP             & 9.9           & 24.2           & 10.1           & 11.3          & 13.9           \\
\ourmethodssmba  & 3.9           & 24.9           & 10.9           & 5.6           & 11.3           \\
\ourmethodmasker & 11.8          & \textbf{25.2}  & \textbf{11.1}  & \textbf{11.6} & \textbf{14.9} \\
\ourmethodmask  & 13.9          & \textbf{25.2}  & 10.2           & 9.9           & 14.8           \\
\ourmethodrand  & \textbf{14.9} & 22.6           & 10.9           & 9.3           & 14.4           \\
\bottomrule
\end{tabular}
\subcaption{Llama-2-7B-Chat.}
\label{tab:llama2_results}
\end{subtable}

%% file: tables/corruption_sensitivity.tex
\centering
\small
\resizebox{0.6\linewidth}{!}{%
\begin{tabular}{@{}llrrrrr@{}}
\toprule
\textbf{Dataset} & \textbf{Method} & $p=.05$ & $p=.10$ & $p=.15$ & $p=.30$ & $p=.50$ \\
\midrule
Companies & \ourmethodrand & 20.2 & 17.7 & 24.4 & 25.7 & 17.4 \\
          & \ourmethodmask & 18.6 & 21.2 & 25.8 & \textbf{27.1} & 19.9 \\
\midrule
Redbook~1 & \ourmethodrand & 39.5 & 37.9 & 39.3 & 39.3 & 34.4 \\
          & \ourmethodmask & 35.3 & 37.4 & \textbf{40.9} & \textbf{40.9} & 38.4 \\
\bottomrule
\end{tabular}%
}
\captionof{table}{Sensitivity to the corruption probability $p$ on Mistral-7B. Scores are LLM-as-a-Judge accuracy ($\times 100$).}
\label{tab:corruption_sensitivity}

%% file: sections/knowledge_instruct_prompts.tex
\subsection{RephraseWeb Prompts for generating synthetic data}
\label{sec:ki-prompts}

For RephraseWeb, we use the prompts proposed by \cite{ovadia2025knowledge} to generate multiple paraphrases of the training data. They use a common system prompt, along with nine different rephrase styles to induce diversity in the generated paraphrases. Both are pasted below. 

%\noindent \textbf{System Prompt}

\begin{tcolorbox}[colback=gray!5, colframe=gray!50, title=System Prompt]
\small
You are an expert in text modification and paraphrasing. Your task:

\begin{itemize}
    \item You will be given an input text (below).
    \item You must produce 1 distinct, long paraphrased version of that text.
\end{itemize}

\noindent\textbf{Requirements:}

\begin{enumerate}
    \item \textbf{Retain Meaning \& Facts:} Each paraphrase must preserve the original text's meaning, factual accuracy, and all specific details. Do not remove, alter, or add any factual information.
    \item \textbf{Variety in Paraphrasing:} Each paraphrase should be substantially different from both the original and from each other in terms of vocabulary, sentence structure, and style.
    \item \textbf{Maintain Length:} Each paraphrase should be approximately the same length as the original text.
    \item \textbf{No Additions or Omissions:} Do not introduce external information or assumptions not present in the original text. Do not remove any facts present in the original.
    \item \textbf{Preserve Specifics:} Do not change proper names, titles, dates, numbers, locations, direct quotes, or other specific references.
    \item \textbf{Clarity and Tone:} Maintain the original clarity and intended tone. The result should read naturally and coherently.
    \item \textbf{Output Format:} Return the paraphrased text directly as a single string. Do not include any preamble or explanation.
    \item \textbf{Include Source:} If the source of the text is mentioned in the input (e.g., title, author, publication, etc.), you must include it in the paraphrased version.
\end{enumerate}
\end{tcolorbox}

Below, we include prompts for generating paraphrases in different styles to increase diversity in the synthetic data.

\begin{tcolorbox}[colback=blue!3, colframe=blue!40, title=Reshuffle, before skip=16pt, after skip=16pt]
\small
Your sub-task is to rearrange the order of the sentences in the text provided below to provide a long paraphrased version. You must significantly change the sequence of sentences, ensuring the final text follows a clear, coherent, and logical structure. While reshuffling, you can make minor modifications to the wording to enhance the flow and coherence of the text. Your output must adhere to the previously mentioned requirements given in the system prompt.
\end{tcolorbox}

\begin{tcolorbox}[colback=blue!3, colframe=blue!40, title=Reword, before skip=16pt, after skip=16pt]
\small
Your sub-task is to rephrase the text provided below, focusing on the choice of words and sentence structure. Your goal is to replace as many words and phrases as possible with synonyms or alternative expressions while maintaining the original meaning, facts, and details. You must ensure that the rephrased text reads naturally and coherently. Your output must adhere to the previously mentioned requirements given in the system prompt.
\end{tcolorbox}

\begin{tcolorbox}[colback=blue!3, colframe=blue!40, title=Restructure, before skip=16pt, after skip=16pt]
\small
Your sub-task is to rewrite the text provided below, focusing on the structure of the sentences. You should rephrase the text by changing the sentence structures, altering the word order, and varying the length of sentences while preserving the original meaning, facts, and details. You must ensure that the rephrased text reads naturally and coherently. Your output must adhere to the previously mentioned requirements given in the system prompt.
\end{tcolorbox}

\begin{tcolorbox}[colback=blue!3, colframe=blue!40, title=Tense Variation, before skip=16pt, after skip=16pt]
\small
Your sub-task is to subtly adjust the verb tenses and aspects in the text provided below. Where appropriate, change some simple past tenses to past perfect, or present tenses to present continuous, while still accurately reflecting the same time frames. Do not introduce new factual content, maintain the same approximate length, and preserve the original meaning and details. Your output must adhere to the previously mentioned requirements given in the system prompt.
\end{tcolorbox}

\begin{tcolorbox}[colback=blue!3, colframe=blue!40, title=Simplify, before skip=16pt, after skip=16pt]
\small
Your task is to simplify the provided text by using more common language, and clearer expressions. You must ensure that the rephrased text reads naturally and coherently without losing any factual information or details. Your output must adhere to the requirements outlined in the system prompt.
\end{tcolorbox}

\begin{tcolorbox}[colback=blue!3, colframe=blue!40, title=Invert, before skip=16pt, after skip=16pt]
\small
Your sub-task is to produce a long paraphrased version of the provided text by inverting its structure. Specifically, you must:

\begin{itemize}
    \item Reverse the overall order of the sentences, starting from the end of the original text and working backward toward the beginning.
    \item While inverting, you may make adjustments to the wording and sentence boundaries to ensure a coherent, logical flow.
    \item Preserve all factual information, names, dates, and other specifics without adding, removing, or distorting any facts.
    \item Maintain the approximate length of the original text and reflect its general tone and clarity.
    \item Ensure the final output reads naturally, as though the text was originally structured in this inverted order.
\end{itemize}

\noindent Your output must adhere to the requirements outlined in the system prompt.
\end{tcolorbox}

\begin{tcolorbox}[colback=blue!3, colframe=blue!40, title=Summarization, before skip=16pt, after skip=16pt]
\small
Your task is to produce a comprehensive and highly detailed knowledge-focused summary of the provided text.

\noindent\textbf{Requirements:}

\begin{enumerate}
    \item \textbf{Thoroughness and Accuracy:} Include all factual information, key points, and essential details, ensuring nothing of importance is omitted.
    \item \textbf{No Alteration of Facts:} The summary must faithfully reflect the original text without adding, removing, or distorting any information.
    \item \textbf{No External Content:} Do not introduce assumptions, opinions, or details not present in the original text.
    \item \textbf{Lexical Variety:} Use a diverse range of vocabulary and phrasing to make the summary more engaging, while maintaining accuracy and avoiding unnecessary repetition.
    \item \textbf{Preserve Tone:} Reflect the general tone and style of the original text.
\end{enumerate}

\noindent Your goal is to create an extensive summary that captures the full breadth of the source material while maintaining clarity, factual integrity, and coherent organization.
\end{tcolorbox}

\begin{tcolorbox}[colback=blue!3, colframe=blue!40, title=Detail Emphasis, before skip=16pt, after skip=16pt]
\small
Your sub-task is to carefully review the provided text and produce a paraphrased version that highlights all minor details and subtle nuances that might otherwise be overlooked.

\noindent\textbf{Requirements:}

\begin{enumerate}
    \item \textbf{Thorough Attention to Detail:} Identify and preserve every small fact, reference, or subtle hint present in the original text, ensuring nothing is lost.
    \item \textbf{No Alteration of Facts:} Accurately reflect every piece of information without distorting or omitting any detail.
    \item \textbf{Clarity and Coherence:} Present the paraphrased text clearly, making sure that the small details fit naturally into the narrative and contribute to overall coherence.
    \item \textbf{Fidelity to Original Tone:} Maintain the original tone, length, and style, incorporating all subtle elements in a way that feels organic and readable.
\end{enumerate}

\noindent Your goal is to produce a paraphrased version that gives as much importance to minor aspects as it does to major points, ensuring full fidelity to the original text.
\end{tcolorbox}

\begin{tcolorbox}[colback=blue!3, colframe=blue!40, title=Middle Restructure, before skip=16pt, after skip=16pt]
\small
Your sub-task is to produce a long paraphrased version of the provided text by reorganizing its structure to begin from its midpoint.

\noindent\textbf{Guidelines:}

\begin{enumerate}
    \item Identify a clear midpoint in the text and start your rewriting from that point.
    \item Move forward through the latter half of the text after this midpoint, maintaining logical flow and coherence.
    \item Once you reach the end of the original text, continue by incorporating the initial portion (the beginning section) at the end, so that the sequence now runs from the middle to the end, and then from the start to the middle.
    \item While restructuring, you may slightly modify wording and sentence boundaries to enhance clarity, coherence, and readability.
    \item Preserve all factual information and details. Do not add or remove any facts. Names, dates, locations, and other specifics must remain accurate.
    \item Reflect the original tone and style.
\end{enumerate}

\noindent Your output must adhere to the requirements outlined in the system prompt.
\end{tcolorbox}

%% file: sections/entigraph_prompts.tex
\subsection{EntiGraph Generation Prompts}
\label{app:entigraph-prompts}

This appendix details the prompts proposed by \cite{yang2025synthetic} for generating synthetic data. 
The pipeline consists of three stages:
(1) entity extraction from source documents,
(2) two-entity relation generation, and
(3) three-entity relation generation.
All prompts are issued to GPT-4o via the Azure OpenAI API.

\subsubsection{Entity Extraction}
\label{app:entity-extraction}

The following system prompt is used to extract salient entities from
each source document. The model is instructed to return structured
JSON.

\begin{tcblisting}{
    colback=gray!5,
    colframe=gray!80,
    title=System Prompt: Entity Extraction,
    listing only,
    breakable,
    listing options={
        basicstyle=\ttfamily\small,
        breaklines=true
    }
}
System:
As a knowledge analyzer, your task is to dissect and understand an
article provided by the user. You are required to perform the
following steps:

1. Summarize the Article:
Provide a concise summary of the entire article, capturing the main
points and themes.

2. Extract Entities:
Identify and list all significant "nouns" or entities mentioned within
the article. These entities should include but are not limited to:

* People:
Any individuals mentioned in the article, using the names or
references provided.

* Places:
Both specific locations and abstract spaces relevant to the content.

* Objects:
Any concrete object that is referenced by the provided content.

* Concepts:
Any significant abstract ideas or themes that are central to the
article's discussion.

Try to exhaust as many entities as possible. Your response should be
structured in JSON format to organize the information effectively.
Ensure that the summary is brief yet comprehensive, and the list of
entities is detailed and accurate.

Use the following response format:

{
  "summary": "<A concise summary of the article>",
  "entities": ["entity1", "entity2", ...]
}
\end{tcblisting}

The user message for entity extraction takes the following form:

\begin{tcblisting}{
    colback=blue!3,
    colframe=blue!40,
    title=User Prompt: Entity Extraction,
    listing only,
    breakable,
    listing options={
        basicstyle=\ttfamily\small,
        breaklines=true
    }
}
Human:
### Document Content:
{document_content}
\end{tcblisting}

\subsubsection{Two-Entity Relation Generation}
\label{app:two-entity-relation}

For each pair of extracted entities $(e_i, e_j)$, the following
system prompt instructs the model to rephrase the document content
with emphasis on each entity and analyze their interaction.

\begin{tcblisting}{
    colback=gray!5,
    colframe=gray!80,
    title=System Prompt: Two-Entity Relation Generation,
    listing only,
    breakable,
    listing options={
        basicstyle=\ttfamily\small,
        breaklines=true
    }
}
System:
You will act as a knowledge analyzer tasked with dissecting an article
provided by the user. Your role involves two main objectives:

1. Rephrasing Content:
The user will identify two specific entities mentioned in the article.
You are required to rephrase the content of the article twice:

* Once, emphasizing the first entity.
* Again, emphasizing the second entity.

2. Analyzing Interactions:
Discuss how the two specified entities interact within the context of
the article.

Your response should clearly separate the rephrased content from the
interaction analysis. Ensure each section includes sufficient context,
ideally referencing the article title to maintain clarity about the
discussion's focus.

Use the following response format:

### Discussion of <title> in relation to <entity1>
<Rephrased content focusing on the first entity>

### Discussion of <title> in relation to <entity2>
<Rephrased content focusing on the second entity>

### Discussion of Interaction between <entity1> and <entity2>
in context of <title>
<Discussion on how the two entities interact within the article>
\end{tcblisting}

The user message for two-entity relation generation takes the
following form:

\begin{tcblisting}{
    colback=blue!3,
    colframe=blue!40,
    title=User Prompt: Two-Entity Relation Generation,
    listing only,
    breakable,
    listing options={
        basicstyle=\ttfamily\small,
        breaklines=true
    }
}
Human:
### Document Content:
{document_content}

### Entities:
- {entity1}
- {entity2}
\end{tcblisting}

\subsubsection{Three-Entity Relation Generation}
\label{app:three-entity-relation}

For each triple of extracted entities $(e_i, e_j, e_k)$, the
following system prompt extends the two-entity setting to three
entities.

\begin{tcblisting}{
    colback=gray!5,
    colframe=gray!80,
    title=System Prompt: Three-Entity Relation Generation,
    listing only,
    breakable,
    listing options={
        basicstyle=\ttfamily\small,
        breaklines=true
    }
}
System:
You will act as a knowledge analyzer tasked with dissecting an article
provided by the user. Your role involves three main objectives:

1. Rephrasing Content:
The user will identify three specific entities mentioned in the
article. You are required to rephrase the content of the article three
times:

* Once, emphasizing the first entity.
* Again, emphasizing the second entity.
* Lastly, emphasizing the third entity.

2. Analyzing Interactions:
Discuss how these three specified entities interact within the context
of the article.

Your response should clearly separate the rephrased content from the
interaction analysis. Ensure each section includes sufficient context,
ideally referencing the article title to maintain clarity about the
discussion's focus.

Use the following response format:

### Discussion of <title> in relation to <entity1>
<Rephrased content focusing on the first entity>

### Discussion of <title> in relation to <entity2>
<Rephrased content focusing on the second entity>

### Discussion of <title> in relation to <entity3>
<Rephrased content focusing on the third entity>

### Discussion of Interaction between <entity1>, <entity2>, and
<entity3> in context of <title>
<Discussion on how the three entities interact within the article>
\end{tcblisting}

The user message for three-entity relation generation takes the
following form:

\begin{tcblisting}{
    colback=blue!3,
    colframe=blue!40,
    title=User Prompt: Three-Entity Relation Generation,
    listing only,
    breakable,
    listing options={
        basicstyle=\ttfamily\small,
        breaklines=true
    }
}
Human:
### Document Content:
{document_content}

### Entities:
- {entity1}
- {entity2}
- {entity3}
\end{tcblisting}